# Maximizing UAV Cellular Connectivity with Reinforcement Learning for BVLoS Path Planning

Mehran Behjati
Research Centre for Human-Machine Collaboration (HUMAC),
Faculty of Engineering & Technology,
Sunway University, Bandar Sunway,
47500, Selangor, Malaysia
mehranb@sunway.edu.my

Rosdiadee Nordin
Future Cities Research Institute,
Sunway University, No. 5, Jalan Universiti, Bandar Sunway 47500,
Selangor, Malaysia
Faculty of Engineering & Technology,
Sunway University, No. 5, Jalan Universiti, Bandar Sunway 47500,
Selangor, Malaysia
rosdiadeen@sunway.edu.my

Nor Fadzilah Abdullah
*Wireless Research@UKM,*
*Department of Electrical, Electronic, and Systems Engineering*
*Universiti Kebangsaan Malaysia*
Bangi, Malaysia
fadzilah.abdullah@ukm.edu.my

***Abstract*—This paper presents a reinforcement learning (RL)-based approach for path planning of cellular-connected unmanned aerial vehicles (UAVs) operating beyond visual line of sight (BVLoS). The objective is to minimize travel distance while maximizing the quality of cellular link connectivity by considering real-world aerial coverage constraints and employing an empirical aerial channel model. The proposed solution employs RL techniques to train an agent, using the quality of communication links between the UAV and base stations (BSs) as the reward function. Simulation results demonstrate the effectiveness of the proposed method in training the agent and generating feasible UAV path plans. The proposed approach addresses the challenges due to limitations in UAV cellular communications, highlighting the need for investigations and considerations in this area. The RL algorithm efficiently identifies optimal paths, ensuring maximum connectivity with ground BSs to ensure safe and reliable BVLoS flight operation. Moreover, the solution can be deployed as an online path planning module that can be integrated into control systems for UAV operations, enhancing their capabilities and safety. The method holds potential for complex long-range UAV applications, advancing the technology in the field of cellular-connected UAV path planning.**

## I. Introduction

Unmanned aerial vehicles (UAVs), commonly referred to as drones, have witnessed rapid adoption across diverse applications such as infrastructure inspection, surveillance, precision agriculture, and package delivery [1, 2]. Their inherent 3D mobility, flexibility, and ability to establish line-of-sight (LoS) communications position UAVs as key enablers in the integration of 5G and emerging 6G networks. Nevertheless, most current UAV operations are restricted to visual line-of-sight (VLoS) flights, whereas the next frontier in UAV technology lies in enabling beyond visual line-of-sight (BVLOS) operations. Achieving BVLOS capabilities critically depends on the availability of reliable communication links to support command-and-control (C&C) functions and high-rate payload transmissions.

Efforts to integrate UAVs into cellular networks generally follow two main paradigms: UAV-assisted wireless communications, where UAVs act as aerial base stations providing coverage in underserved regions, and cellular-connected UAVs, where UAVs function as aerial user equipment (UE) directly leveraging terrestrial cellular infrastructure [3]. The latter approach offers cost-effective, wide-area connectivity without the need for dedicated infrastructure. However, terrestrial cellular networks are inherently optimized for ground users, leading to non-trivial challenges when extended to aerial scenarios, including antenna downtilt, increased interference, and coverage gaps at higher altitudes [4]. Addressing these issues to guarantee seamless and resilient UAV connectivity has thus become a central research direction for beyond 5G and 6G wireless systems [5, 6].

Path planning is a fundamental component of UAV operations, as it determines the route from a starting point to a designated destination, balancing mission objectives with safety and efficiency requirements. Existing approaches can broadly be divided into global (offline) methods, which compute an optimal trajectory in a known environment, and local (online) methods, which adaptively plan paths in dynamic or partially unknown environments through onboard sensing and real-time decision-making [7]. The choice between global and local strategies depends on factors such as computational complexity, environmental predictability, and mission-critical constraints.

To address the complexity of UAV path planning, numerous optimization and bio-inspired algorithms have been explored, including ant colony optimization, genetic algorithms, particle swarm optimization, and whale optimization [8]. While these methods are effective in certain scenarios, they often suffer from premature convergence and sub-optimal performance in high-dimensional or dynamic environments. More recently, reinforcement learning (RL) has gained traction as a powerful alternative, enabling UAVs to learn adaptive path planning strategies through interaction with the environment. Classical RL methods such as Q-learning have shown promising results in dynamic environments [9, 10], while recent studies on cellular-connected UAVs demonstrate the use of RL integrated with A* algorithms [11], deep RL for 3D trajectory optimization [12], and RL-based designs for minimizing mission time while ensuring strong connectivity [13]. These works confirm the feasibility of RL-driven UAV navigation but also highlight the need for further exploration.

Despite these advances, important research gaps remain. First, limited studies explicitly address the dual objective of UAV path planning under both mobility and cellular connectivity constraints, particularly in the context of BVLOS safety requirements. Second, many existing approaches rely on idealized network models, whereas realistic empirical constraints of terrestrial cellular networks—such as limited aerial coverage and interference dynamics—are insufficiently considered. These limitations underscore the importance of developing optimization strategies that explicitly incorporate communication-aware objectives within the path planning process.

This paper seeks to address these gaps by proposing a RL-based approach to optimize UAV path planning with simultaneous consideration of travel efficiency and cellular connectivity. Specifically, the proposed framework employs RL to train UAV agents in a grid-based aerial environment modeled with empirical coverage characteristics, thereby producing trajectories that ensure reliable communication with terrestrial base stations.

The main contributions of this paper are summarized as follows:

- Proposing an RL-based technique for UAV path planning that explicitly accounts for aerial coverage constraints imposed by terrestrial cellular networks.
- Developing a grid map-based environment model and applying Q-learning to train the UAV agent and derive optimal navigation policies.
- Evaluating the effectiveness of the proposed approach through simulation experiments, with comparative analysis against benchmark RL algorithms.
- Discussing the implications of cellular connectivity-aware path planning for UAV safety and reliability in BVLOS operations.

The remainder of this paper is organized as follows. Section II details the proposed methodology, Section III presents simulation results and discussion, and Section IV concludes the paper with perspectives on future research directions.

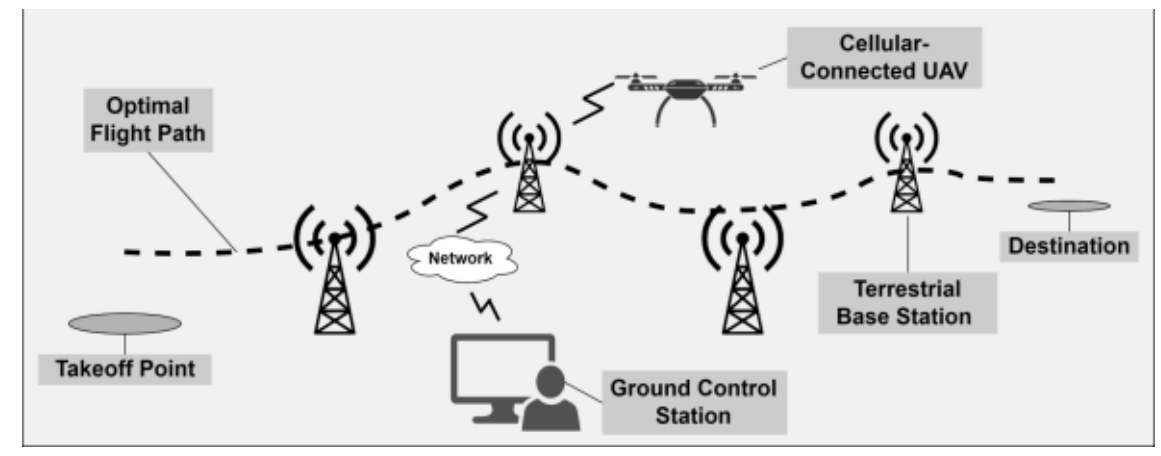

Fig. 1. Schematic diagram of the considered path planning problem

## II. PROBLEM FORMULATION AND SYSTEM MODEL

The problem of UAV path planning in cellular-connected environments can be formulated as a sequential decision-making task that accounts for both mobility efficiency and wireless connectivity quality. Specifically, a UAV is required to travel from a defined takeoff point to a target destination while maintaining robust connectivity with terrestrial base stations (BSs) to ensure safe BVLoS operations. Figure 1 illustrates the scenario where multiple BSs are spatially distributed across the operational area, providing aerial coverage to the UAV throughout its mission.

### *A. Mission Objective*

The main objective is twofold:

1. Minimize travel distance, ensuring efficient utilization of limited onboard energy resources.
2. Maximize connectivity reliability, maintaining continuous C&C and payload data transmissions with ground control systems (GCS).

These dual objectives are particularly important in BVLoS scenarios, where communication blackouts may compromise safety and mission success. Thus, the UAV must select a path that simultaneously balances flight efficiency and signal quality.

### *B. Markov Decision Process Formulation*

The UAV path planning problem is modeled as a Markov Decision Process (MDP), defined by the tuple ($S$, $A$, $P$, $R$, $\gamma$)where:

- $S$ represents the state space, describing the UAV's current position, movement parameters, and communication conditions.
- $A$ denotes the action space, i.e., the set of possible trajectory decisions at each step.
- $P$ defines the state transition probabilities under a chosen action.
- $R$ is the reward function, designed to encourage distance minimization while ensuring connectivity.
- $\gamma$ is the discount factor balancing immediate and future rewards.

The UAV agent seeks to identify an optimal policy $\pi*(s)$ that maximizes the expected cumulative reward across its trajectory.

### *C. Aerial Coverage Constraints*

Realistic modeling of aerial coverage is critical to capture the limitations of terrestrial cellular networks. According to empirical studies [14], existing 4G BSs can provide reliable coverage up to 500 m radius and 85 m altitude in suburban environments. This highlights two constraints:

- UAVs flying outside these thresholds face reduced signal strength (RSRP) and degraded link quality (RSRQ).
- High mobility and altitude variations exacerbate interference due to side-lobe reception from downtilted BS antennas.

By embedding these empirical constraints, the path planning framework reflects realistic aerial network behavior, improving its practical applicability.

### *D. Environment Model*

The operational space is discretized into a grid-based environment, where each grid cell corresponds to a 250 m² area. This abstraction enables efficient mapping of UAV positions to coverage zones and allows the agent to perceive whether a state lies within or outside reliable cellular connectivity. The grid representation also facilitates scalable training, as the UAV interacts with the environment over discrete state-action transitions.

In summary, the system model establishes UAV path planning as an MDP constrained by terrestrial cellular coverage characteristics. This formulation provides a foundation for employing reinforcement learning techniques to derive adaptive, connectivity-aware flight trajectories.

## III. REINFORCEMENT LEARNING FRAMEWORK FOR UAV PATH PLANNING

Building upon the problem formulation, this section describes the RL framework employed to optimize UAV path planning. RL is particularly suited to this problem, as it allows the UAV agent to learn adaptive navigation strategies through interaction with its environment, without requiring an explicit model of the environment's dynamics.

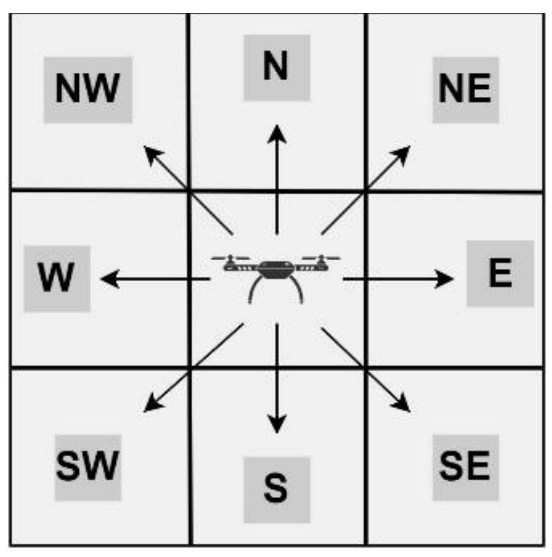

Fig. 2. Considered actions

### A. *Learning Algorithm: Q-Learning and SARSA*

The core RL algorithm employed is Q-learning, a model-free method that iteratively learns an action-value function Q(s,a)Q(s,a)Q(s,a), estimating the expected cumulative reward of taking action aaa in state sss. The update rule is expressed as [15]:

$$Q(s_t, a_t) \leftarrow Q(s_t, a_t) + \alpha[r_{t+1} + \gamma \max_a Q(s_{t+1}, a_t) - Q(s_t, a_t)] \quad (1)$$

where, $s_t$ is the current state, $a_t$ is the current action, $a_{t+1}$ is the subsequent action, $s_{t+1}$ is the state after taking action $a_t$, $r_{t+1}$ is the reward, $\alpha$ is the learning rate, and $\gamma$ is the discount factor.

The pseudocode for the Q-Learning algorithm is as follows:

1. Algorithm parameters: step size $\alpha \epsilon (0,1]$, $\epsilon > 0$
2. Initialize $Q(s, a)$, for all $s \in S^+, a \in A(s)$ arbitrary except that $Q(termination, .) = 0$
3. Loop for each episode
4. Initialize S
5. Loop for each step of the episode:
6. Choose $A$ from $S$ using a policy derived from $Q$
7. Take action $A$, observe $R, S'$
8. $Q(s_t, a_t) \leftarrow Q(s_t, a_t) + \alpha[r_{t+1} + \gamma \max_a Q(s_{t+1}, a_t) - Q(s_t, a_t)]$ <br> $S \leftarrow S'$
9. Until S is terminal

The Q-learning algorithm updates the Q-values based on the observed rewards and actions, allowing the UAV agent to learn optimal policies for path planning in the cellular-connected environment.

To compare the performance of the developed approach under different algorithms, the SARSA algorithm is also utilized. The Q-function update equation for SARSA is given by [15]:

$$Q(s_t, a_t) \leftarrow Q(s_t, a_t) + \alpha[r_{t+1} + \gamma Q(s_{t+1}, a_{t+1}) - Q(s_t, a_t)]. \quad (2)$$

Equation (2) updates the Q-value based on the reward received, the next state, and the next action taken in the SARSA algorithm. By considering different action selections and updates, SARSA provides an alternative perspective on the learning process and allows for a comparison of its performance with Q-learning.

### B. *State Space Definition*

The state representation includes:

- UAV's current grid position.
- Movement direction and speed.
- Signal strength (RSRP/RSRQ) of nearby BSs.

This multi-dimensional state space captures both navigation context and communication reliability, enabling the agent to make connectivity-aware decisions.

### C. *Action Space Definition*

The action space encompasses the possible UAV movements based on received cellular signal strength from nearby base stations. By considering signal power levels, the agent navigates efficiently to maximize communication quality and minimize travel distance. Figure 2 shows the considered actions, allowing the UAV to adapt its trajectory dynamically based on real-time signal feedback.

### D. *Reward Function*

The reward function plays a critical role in guiding the learning process of the agent. It provides feedback to the UAV on the quality of its decisions. In this study, the reward function is designed to encourage the UAV to maximize the received signal quality of nearby cellular towers while minimizing the distance to the destination. The consideration of aerial signal quality is based on the empirical reference signal received power (RSRP) and reference signal received quality (RSRQ) presented in [14]. The reward function also penalizes the UAV for violating constraints. Table 1 presents the setting of the reward function considered in this paper.

TABLE I. SETTINGS OF THE REWARD FUNCTION

| State | Reward |
| --- | --- |
| UAV reaches the destination | +10 |
| UAV moves to a valid state | -1 |
| UAV moves to an invalid state | -10 |
| UAV moves in reliable zone | -0.3 |

### E. *Training the Agent:*

The UAV agent is trained using the Q-learning algorithm in the defined environment. Initially, the agent has no prior knowledge of the environment and randomly selects actions. As the training progresses, the agent learns from the rewards obtained by performing actions and updates its Q-values accordingly. The exploration-exploitation trade-off is achieved using an ε-greedy policy, where the agent selects the action with the maximum Q-value, $Q_t(a)$, with a probability of (1-ε) and a random action, $a_i \ \forall \ i \in \{1, \ldots, k\}$, with a probability of ε, as follows [15]:

$$a_t \leftarrow \begin{cases} \arg\max_a Q_t(a) & \text{with probability } 1 - \varepsilon \\ a \sim Uniform(\{a_1, \ldots, a_k\}) & \text{with probability } \varepsilon \end{cases} \quad (3)$$

### F. *Testing and Evaluation:*

After the agent is trained, it is tested and evaluated in a simulation environment. The performance of the agent is measured based on metrics such as the distance travelled and the signal quality of nearby cellular towers. The simulation results provide insights into the effectiveness of the proposed approach in optimizing UAV path planning.

## *G. Policy Improvement:*

Based on the testing and evaluation results, the policy can be refined to improve the performance of the agent. The policy improvement iteration allows for the determination of an improved policy and the calculation of its state-action value function. This iterative process continues until an optimal policy is reached.

By following this methodology, the proposed RL-based technique optimizes the path planning for UAVs in a cellular-connected environment. The integration of Q-learning with the grid map environment and the defined reward function enables the agent to make informed decisions, minimizing the travelling distance and maximizing the quality of wireless connectivity provided by terrestrial cellular networks.

# IV. Results and Performance Analysis

The performance of the proposed RL-based UAV path planning framework was validated through simulations conducted in the MATLAB environment. The simulation scenario was designed to reflect a 2D task space of 10 km × 4 km, representing a typical suburban environment where UAVs perform BVLoS missions. Terrestrial BSs were modeled as grid-based coverage zones, each approximating the empirical aerial connectivity characteristics reported in [14].

## *A. Convergence Behavior of Learning Algorithms*

To evaluate the efficiency of the learning process, the cumulative rewards obtained during training were monitored. Figure 3 illustrates the convergence curves of the Q-learning and SARSA algorithms across multiple training episodes. Both algorithms demonstrate convergence, but the Q-learning approach achieves approximately 10% higher cumulative rewards, indicating superior learning efficiency. This improvement highlights Q-learning's ability to exploit the action-value maximization principle, enabling the UAV agent to discover more favorable policies with fewer episodes.

From a BVLoS safety perspective, faster convergence translates into reduced training overheads and quicker deployment of reliable flight strategies. Such convergence advantages become increasingly significant in real-time or large-scale multi-UAV deployments, where computational and energy budgets are constrained.

## *B. Path Optimality and Connectivity Reliability*

The trajectories planned by Q-learning and SARSA are depicted in Figure 4. The green circular regions represent reliable coverage zones, approximating areas where aerial link quality remains above the empirical RSRP/RSRQ thresholds [14]. Both algorithms successfully guide the UAV to the target destination; however, clear differences emerge:

- Q-learning path length: 52 steps (≈ 13 km).
- SARSA path length: 57 steps (≈ 14.25 km).

Thus, Q-learning achieves a 9% reduction in travelled distance, reflecting more energy-efficient flight. Table II summarizes the performance comparison, showing that Q-learning also yields higher coverage reliability (78% vs. 71%), meaning a greater proportion of the UAV's trajectory is executed under strong cellular connectivity.

$$Coverage\ reliability = \frac{L_{reliable}}{L_{total}} \times 100\%, \tag{4}$$

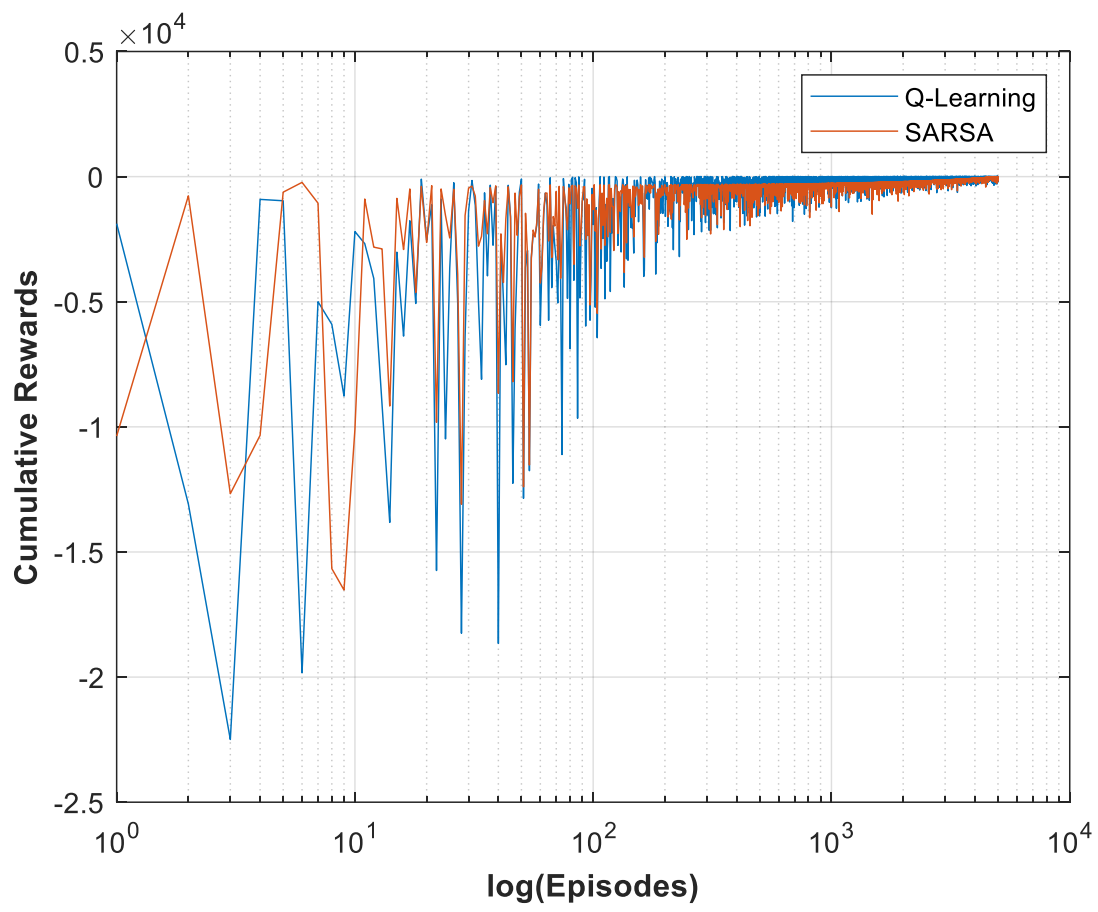

Fig. 3. Performance Comparison: Cumulative Rewards of Q-learning and SARSA Algorithms

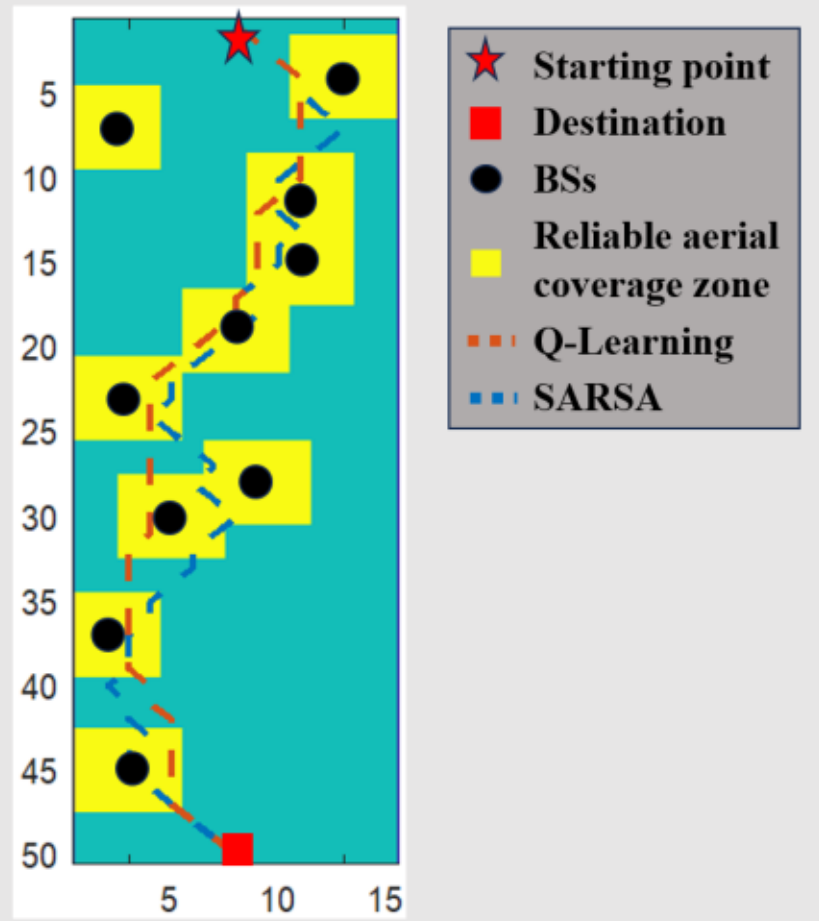

Fig. 4. Comparison of Optimal Paths: Q-learning vs. SARSA Algorithm

where $L_{reliable}$ represents the path length within the reliable aerial coverage zone and $L_{total}$ is the total path length. These results confirm that Q-learning is better at balancing the trade-off between minimizing distance and maintaining reliable connectivity. This dual optimization is critical for BVLoS flights, where energy efficiency and link availability directly affect mission success and flight safety.

## *C. Discussion of Algorithmic Insights*

The observed superiority of Q-learning can be attributed to its off-policy nature, which allows the agent to evaluate and update state-action values based on the best possible actions, independent of the current behavior policy. In contrast, SARSA's on-policy updating ties its performance to the agent's exploratory actions, leading to more conservative learning. While this conservatism may reduce the risk of unsafe exploration in certain cases, it also results in longer trajectories and lower connectivity reliability.

An additional observation is that the reward shaping plays a decisive role in performance, particularly the balance between distance penalties and connectivity incentives. The inclusion of connectivity-based penalties (-0.3) ensures that the UAV avoids marginal coverage regions, directly enhancing communication reliability. Fine-tuning such weights could further improve trade-offs between energy consumption and link quality, especially for longer missions.

### D. *Limitations and Extensions*

While the presented results demonstrate the efficacy of Q-learning, several limitations remain:

- Static environment assumption: The simulations assume fixed BS placements and static coverage zones. In practice, interference fluctuations, user load, and environmental factors may alter aerial connectivity.
- Single-agent setting: Only one UAV is considered. In real-world deployments, multi-UAV cooperation (e.g., coordinated coverage-aware swarm missions) may introduce new challenges in collision avoidance and spectrum sharing.
- Energy modeling: Although path length serves as a proxy for energy consumption, incorporating battery discharge models and propulsion energy dynamics would provide more realistic insights.

Future extensions could integrate deep reinforcement learning (e.g., DQN, PPO) to scale performance in high-dimensional environments, or incorporate federated learning frameworks to enable distributed policy training across multiple UAVs and environments. Furthermore, additional metrics—such as energy efficiency (J/km), handover rate, and signal outage probability—should be explored to better capture UAV connectivity performance in real-world scenarios.

In summary, the results demonstrate that the proposed RL-based framework is effective in generating connectivity-aware UAV trajectories. Q-learning consistently outperforms SARSA by delivering shorter paths, higher cumulative rewards, and improved coverage reliability. These findings provide evidence that reinforcement learning can serve as a viable tool for cellular-connected UAV path planning, supporting the safe and reliable realization of BVLoS operations.

## V. Conclusion

This paper presented a reinforcement learning-based technique for optimizing UAV path planning in cellular-connected environments. By formulating the problem as an MDP and incorporating empirical aerial coverage constraints, the proposed framework addresses both energy efficiency and connectivity reliability—two critical requirements for BVLoS flight safety. The Q-learning algorithm consistently outperformed SARSA in simulations, achieving shorter trajectories, higher convergence rewards, and greater coverage reliability. These results confirm that connectivity-aware RL can effectively guide UAVs in balancing mission efficiency with robust command-and-control (C&C) link availability.

Beyond the demonstrated improvements, the study underscores several avenues for future research. First, incorporating more realistic energy consumption models and handover dynamics would enhance practical applicability. Second, extending the framework to multi-UAV scenarios could enable cooperative and interference-aware swarm missions. Third, advanced RL paradigms such as deep Q-networks (DQN), policy optimization (PPO), and federated learning may be explored to scale performance in complex, dynamic, and large-scale environments.

In conclusion, the proposed Q-learning-based path planning framework contributes to the development of safe, reliable, and efficient BVLoS UAV operations, and represents a step forward in integrating UAVs into next-generation 5G and 6G cellular networks.